# The Clustering of Author's Texts of English Fiction in the Vector Space of Semantic Fields

Bohdan Pavlyshenko

*Ivan Franko Lviv National University,Ukraine,  pavlsh@yahoo.com*

**Abstract**

The clustering of text documents in the vector space of semantic fields and in the semantic space with orthogonal basis has been analysed. It is shown that using the vector space model with the basis of semantic fields is effective in the cluster analysis algorithms of author's texts in English fiction. The analysis of the author's texts distribution in cluster structure showed the presence of the areas of semantic space that represent the author's ideolects of individual authors. SVD factorization of the semantic fields matrix makes it possible to reduce significantly the dimension of the semantic space in the cluster analysis of author's texts.

**Introduction**

In the analysis of author's texts it is effective to use the methods of data mining, including clustering methods. In text arrays clusterization a vector model of text documents is used, according to which the documents are considered as vectors in some vector space, formed by quantitative characteristics of words [Pantel and Turney 2010]. As quantitative characteristics the frequencies of keywords are widely used. One of the problems of such an approach is caused by a great dimension of text documents space, which is caused by the size of the vocabulary of text array under analysis. A promising trend is the use of vector space with a basis formed by quantitative characteristics of word associations, in particular semantic fields. A semantic field is considered as a set of words that are united under some common concept. The examples of semantic fields can be the field of motion, the field of communication, the field of perception, etc. The number of semantic fields is significantly smaller than the size of a word dictionary, and it reduces the amount of necessary calculations. Similar objects are semantic networks that describe the relationships among different concepts. An example of lexicographic computer system, which represents the semantic network of links between words, is a WordNet system, developed at Princeton University [Fellbaum 1998]. This system is based on an expert lexicographic analysis of semantic structural relationships that describe the denotative and connotative characteristics of dictionary word composition. The paper [Gliozzo and Strapparava 2009] considered the concept of semantic domain, which describes certain semantic areas of various issues discussed, such as economics, politics, physics, programming, etc. The algorithms of clusterization and classification are often used in data mining [Sebastiani 2002; Manning, Raghavan and Schütze 2008]. Recording the text semantics in the problems of text documents clustering makes it possible to obtain the clustering of greater accuracy [Shehata, Karray and Kamel 2006]. In [Larsen and Aone, 1999] text clustering algorithms and the evaluation of their effectiveness are described.

Let us consider the vector model of text documents in the space of semantic and thematic fields in terms of the use of this model in agglomerative clustering algorithms. We apply the singular spectrum of the matrices of semantic and thematic fields of text documents to form an orthogonal semantic space. We perform an experimental analysis of author's texts in English fiction using clustering algorithms in the space of semantic fields and in the semantic space with orthogonal basis.

**The model of text documents clustering in the space of semantic fields**

Let us consider a model based on a set theory, which describes a set of text documents, word composition, and semantic fields. A set of text documents we describe as:

$$D = \{d_j \mid j = 0,1,2...,N_d\}. \tag{1}$$

Let us introduce a set of semantic fields

$$S = \{s_k \mid k = 1,2...,N_s\}. \tag{2}$$

On the basis of word composition of semantic fields we form a matrix of a feature-document type where the features are the frequencies of semantic fields in the documents:

$$M_{sd} = \left(p_{kj}^{sd}\right)_{k=1, j=1}^{N_s, N_d}. \tag{3}$$

The frequencies of semantic fields $p_{kj}^{sd}$ are defined as the sums of word text frequencies that are included into these semantic fields. The values of these frequencies are normalized so that their sum for each document is equal to 1. The vector

$$V_j^s = \left(p_{1j}^{sd}, p_{2j}^{sd},...,p_{N_s j}^{sd}\right)^T \tag{4}$$

displays the document $d_j$ in $N_s$-dimensional space of text documents. The introduction of the semantic fields space not only reduces the size of the problem of texts analysis, but also introduces a new basis for text descriptions. In the semantic basis qualitatively new clustering text documents can be observed.

Let us consider the documents groupings by semantic features using hierarchical clustering algorithm. Suppose there is a set of text documents $D$, which is described by the expression (1) and a set of clusters

$$C = \{c_m \mid m = 0,1,2...,N_c\}. \tag{5}$$

It is necessary to build a mapping of documents set by clusters set

$$U_{DC} : D \to C. \tag{6}$$

The mapping $U_{DC}$ specifies the data model, which is a solution of clustering problem. Each element $C_m$ of the set of clusters $C$ consists of a subset of text documents that are similar to each other according to some quantitative similarity measure $r$

$$c_m = \{d_i, d_j \mid d_i \in D, d_j \in D, r(d_i, d_j) < \varepsilon\}, \tag{7}$$

where $\varepsilon$ defines some threshold for including the documents into the cluster. The value $r(d_i,d_j)$ is the distance between the elements $d_i$ and $d_j$ and if it is less than some value, then the sample elements are considered as being similar and belonging to a common cluster. Since the concept of distance is introduced on the set of text documents, each document is represented as a point in $N_s$-dimensional space of $R^{N_s}$ semantic features. In our studies we calculate the Euclidean distance. Let us consider a hierarchical agglomerative clustering method. At the first step the entire set of text documents is considered as a set of clusters. At the next step two documents, close to each other, are combined into one common cluster, a new set at this step is composed of $N_d$-1 clusters. Reiterating the steps at which the clusters will be combined, we obtain a set of $N_c$ clusters. The process of combining the clusters ends at that step of the algorithm, when no pair of clusters meets the threshold of combining for the proximity measure of elements. There are different methods of forming and joining clusters on the basis of distances between the objects within the cluster. One of effective methods of text documents clustering in the semantic fields space is a Ward's method. This method calculates the squares of Euclidean distances from individual documents to the center of each cluster. Then these distances are being summed. Only those clusters can be combined in a new one, the combination of which gives the smallest increase in the sum of squares of these distances. The graphic representation of the hierarchical clustering result is a dendrogram that indicates the process of agglomerative clustering aggregation. The numbers of clusters are on the abscissa axis, and the distances between clusters are on the ordinate axis. At certain values of the distances the clusters begin to merge. With the increase of intercluster distance the clusters are merging up to complete union of clusters into one cluster. Therefore, in order to obtain informative cluster structure, some threshold of intercluster distance must be chosen, when the optimal cluster structure, from the point of view of the text arrays analysis, is formed.

**Text Analysis in the Semantic Space with Orthonormal Basis**

The method of latent-semantic analysis, based on the singular decomposition of keywords frequencies matrix, allows to reduce significantly the dimension of the documents vector space [Deerwester at al. 1990]. Let us consider the singular decomposition of the semantic fields frequencies matrix. Let there is a matrix of a "semantic_fields_frequencies-documents" type $M_{sd}$, which is described by the formula (12). The vector $V_j^s$ (14) displays the document $d_j$ in $N_s$-dimensional space of text documents. The product of two vectors $(V_p^s)^T V_q^s$ determines the quantitative measure of similarity of these vectors in $N_s$-dimensional semantic space of text documents. Accordingly, the product of two matrices $(M_{sd})^T M_{sd}$ contains scalar products of vectors $(V_p^s)^T V_q^s$ of all documents and it reflects their correlations in semantic vectors space. The singular matrix decomposition $M_{sd}$ looks as

$$M_{sd} = U_{sd} \Sigma_{sd} Y_{sd}^T \quad (8)$$

The diagonal matrix $\Sigma_{sd}$ contains singular numbers in descending order. If we take the $K$ of the largest singular numbers of the matrix $\Sigma_{sd}$ and, correspondingly, the $K$ of singular vectors of the matrices $U_{sd}$ and $Y_{sd}$, we will get the $K$-rank approximation of the matrix $M_{sd}$:

$$(M_{sd})_K = (U_{sd})_K (\Sigma_{sd})_K (Y_{sd})_K^T \quad (9)$$

The matrix $(Y_{sd})_K$ reflects the relations between the vectors of the documents $\hat{V}_j^s$ in the new combined $K$-dimensional orthogonal semantic space. The relations between the vector $V_j^s$ of the document in the original semantic space and the vector $\hat{V}_j^s$ in orthogonal semantic space can be described as

$$\begin{aligned} V_j^s &= (U_{sd})_K (\Sigma_{sd})_K \hat{V}_j^s \\ \hat{V}_j^s &= (\Sigma_{sd})_K^{-1} (U_{sd})_K^T V_j^s \end{aligned} \quad (10)$$

Apparently, the number $K$ can be smaller significantly than the $N_s$ dimension of the initial semantic space. This reduces the dimension of the problem of the analysis of text documents similarity in the semantic vector space.

**Experimental Part**

For the experimental study of text documents clustering in the space of semantic fields we chose a text base containing 503 literary works of 17 authors. For the semantic space generation we chose the words grouped by the semantic fields of nouns and verbs in the semantic network WordNet [Fellbaum 1998]. The semantic fields in the WordNet network (http://wordnet.princeton.edu) are represented as lexicographic files. In our studies we have used the semantic fields of nouns and verbs. The semantic fields of nouns consist of 26 lexicographic files out of which we have selected 54464 words. The semantic fields of verbs contain 15 lexicographic files out of which we have selected 9097 words. The derivative forms of words were also included into the semantic fields. Lexicographic files WordNet for nouns and verbs have the names that define the semantic core of these fields: noun.tops, noun.act, oun.animal, noun.artifact, noun.attribute, noun.body, noun.cognition, noun.communication, noun . event, noun.feeling, noun.food, noun.group, noun.location, noun.motive, noun.object, noun.person, noun.phenomenon, noun.plant, noun.possession, noun.process, noun.quantity , noun.relation, noun.shape, noun.state, noun.substance, noun.time, verb.body, verb.change, verb.cognition, verb.communication, verb.competition, verb.consumption, verb.contact, verb.creation, verb.emotion, verb.motion, verb.perception, verb.possession, verb.social, verb.stative, verb.weather. We have selected the agglomerative clustering method with Euclidean intercluster distance. For the formation of clusters we have chosen the Ward's method. Figure 1 shows the clusters dendrogram, which describes the formation of cluster structure. This dendrogram represents the formation of the first 20 clusters. The clustering process is stopped as soon as the cluster structure contains 20 clusters. Figure 2 shows the histograms of the texts authors distribution in the clusters. Each histogram corresponds to a particular cluster. The number of the column indicates the corresponding number of the text author. These histograms reflect how the documents of different groups are distributed in each cluster. There are the clusters where the texts of separate authors stand on the dominant position. As follows from the data given, some clusters contain the texts of wide semantic spectrum. Obviously, the area of these clusters in the semantic space is semantically homogeneous and it has semantically low differentiating potential. However, there are also such clusters where the texts of one or more authors stand on the dominant position. Such clusters characterize the author's idiolect of individual authors. The semantic space areas of these clusters have differentiating potential for author's idiolect and they can be used while analyzing authors' texts as an additional factor in the analysis of author's lexicon. The areas of the semantic space, corresponding to the clusters, where two or more authors dominate, can be considered as the fields of the semantic similarity of these authors.

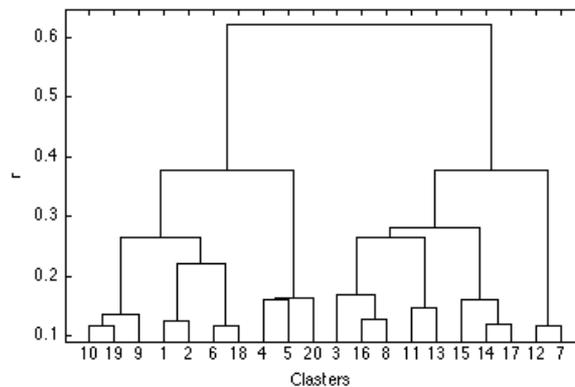

*Fig.1 The dendrogram of hierarchical clustering of author's texts in the space of semantic fields.*

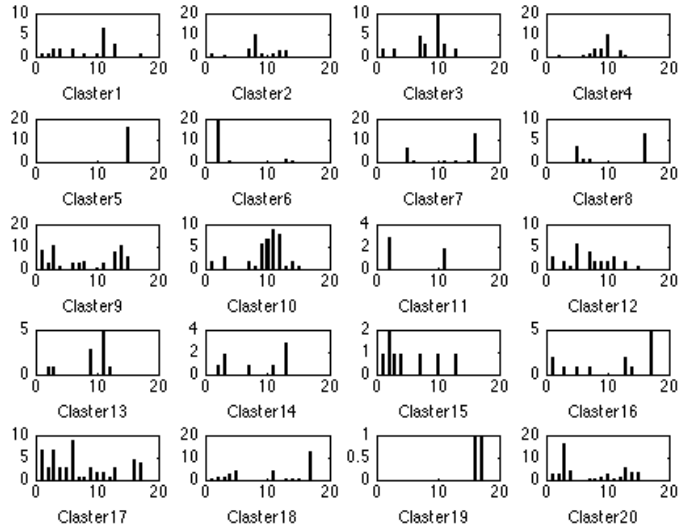

*Fig. 2 The distribution of the texts authors by the clusters in the space of semantic fields.*

Our next step is to consider the clustering of author's texts in orthogonal low-dimensional space of secondary semantic fields, generated by SVD factorization of the semantic fields matrix. Figure 3 shows the first 10 singular values of the semantic frequencies matrix $M_{sd}$. A significant decreasing of the values of singular numbers is being observed. For the formation of orthogonal semantic subspace we have taken the coordinates of secondary semantic fields that correspond to the first 10 singular numbers of the matrix $M_{sd}$. On the basis of generated low-dimensional orthogonal space we have conducted similar calculations of the dendrogram (Fig. 4) and the distribution of authors in clusters (Fig. 5). As follows from the data obtained, the clusters with the predominance of individual authors are also present. These clusters also characterize the semantic area of author's idiolect in low-dimensional semantic space with orthogonal basis. We have also conducted the studies for the orthogonal subspace with the dimension of 3. In this case the clusters, where the texts of certain author dominate, are not observed.

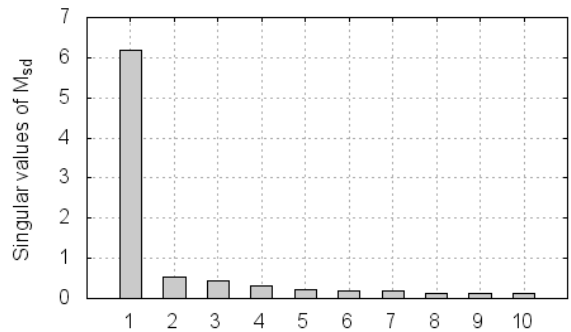

*Fig.3 Singular values for the matrix of semantic fields.*

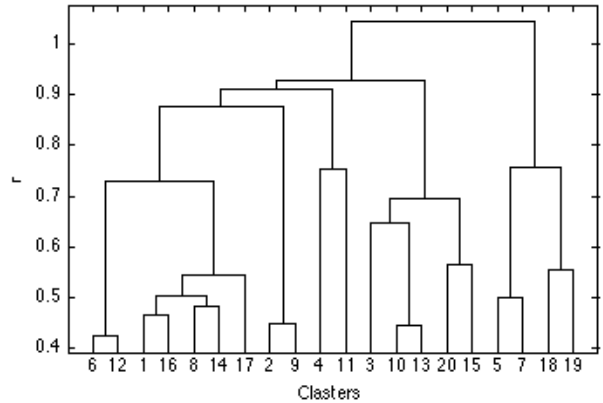

*Fig.4 The dendrogram of hierarchical clustering of author's texts
in the semantic space with orthogonal basis.*

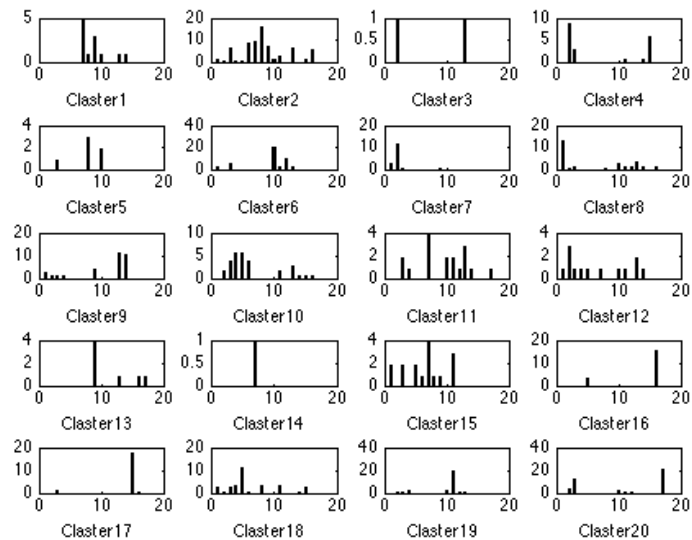

*Fig.5 The distribution of the authors of texts by clusters
in the semantic space with orthogonal basis.*

**Summary and Conclusions**

Considered in the paper is the clustering of text documents in the vector space of semantic fields and in the semantic space with orthogonal basis. The dimension of the vector space basis of semantic fields is significantly lower in comparison with clustering methods by keywords. Orthogonal semantic basis is formed on the basis of SVD factorization of the semantic fields matrix of the text documents. Using the vector space model with the basis of semantic fields is effective in the cluster analysis algorithms of author's texts in English fiction. The frequency characteristics of semantic fields were considered as semantic features. The analysis of the author's texts distribution in cluster structure showed the presence of the areas of semantic space that represent the author's lexicon of individual authors. The clustering of author's texts in the space of semantic fields allows to detect the semantic areas of author's idiolect that are identified by the clusters with dominant text authors. SVD factorization of the semantic fields matrix makes it possible to reduce significantly the dimension of the semantic space in the cluster analysis of author's texts.